%% file: main.tex
\documentclass[journal]{IEEEtran}

\usepackage{multirow}
\usepackage{cite}

\ifCLASSINFOpdf
  \usepackage[pdftex]{graphicx}
\else
  \usepackage[dvips]{graphicx}
\fi
\usepackage{amsmath}
\usepackage{algorithmic}

\usepackage{array}

\usepackage{changepage}
\usepackage{subcaption}

\usepackage[pagebackref=true,breaklinks=true,letterpaper=true,colorlinks,bookmarks=false]{hyperref}

\providecommand{\etal}{\textit{et al.}}

\begin{document}

\title{$\mathcal{R}^2$-CNN: Fast Tiny Object Detection in Large-scale Remote Sensing Images}

%

\author{Jiangmiao Pang,
        Cong Li\thanks{J. Pang, Z. Xu, H. Feng are with Zhejiang University, Hangzhou, China (e-mails: pjm@zju.edu.cn, xuzh@zju.edu.cn, fenghj@zju.edu.cn).},
        Jianping Shi\thanks{C. Li and J. Shi are with SenseTime Research, Beijing, China (e-mails: licong@sensetime.com, shijianping@sensetime.com).},
        Zhihai Xu\thanks{Digital Object Identifier 10.1109/TGRS.2019.2899955},
        Huajun Feng
}

%
%

\markboth{IEEE TRANSACTIONS ON GEOSCIENCE AND REMOTE SENSING}%
{Shell \MakeLowercase{\textit{et al.}}: Bare Demo of IEEEtran.cls for IEEE Journals}
%



\maketitle

\input{sections/abstract}

%
\IEEEpeerreviewmaketitle

%
%
%
%
\input{sections/introduction}
\input{sections/related_work}
\input{sections/methodology}

\input{sections/experiments}
\input{sections/conclusion}


%

%
%
%
%
%

\ifCLASSOPTIONcaptionsoff
  \newpage
\fi



%


\bibliographystyle{IEEEtran}
\bibliography{references}

%



\end{document}

%% file: sections/abstract.tex

\begin{abstract}
Recently, the convolutional neural network has brought impressive improvements for object detection.
However, detecting tiny objects in large-scale remote sensing images still remains challenging.
First, the extreme large input size makes the existing object detection solutions too slow for practical use.
Second, the massive and complex backgrounds cause serious false alarms.
Moreover, the ultratiny objects increase the difficulty of accurate detection.
To tackle these problems, we propose a unified and self-reinforced network called remote sensing region-based convolutional neural network ($\mathcal{R}^2$-CNN),
composing of backbone Tiny-Net, intermediate global attention block, and final classifier and detector.
Tiny-Net is a lightweight residual structure, which enables fast and powerful features extraction from inputs.
Global attention block is built upon Tiny-Net to inhibit false positives.
Classifier is then used to predict the existence of targets in each patch, and detector is followed to locate them accurately if available.
The classifier and detector are mutually reinforced with end-to-end training, which further speed up the process and avoid false alarms.
Effectiveness of $\mathcal{R}^2$-CNN is validated on hundreds of GF-1 images and GF-2 images that are 18 000 $\times$ 18 192 pixels, 2.0-m resolution, and 27 620 $\times$ 29 200 pixels, 0.8-m resolution, respectively.
Specifically, we can process a GF-1 image in 29.4 s on Titian X just with single thread.
According to our knowledge, no previous solution can detect the tiny object on such huge remote sensing images gracefully.
We believe that it is a significant step toward practical real-time remote sensing systems.
\end{abstract}


%% file: sections/introduction.tex
\section{Introduction}
\IEEEPARstart{T}{hanks} to the development of optical remote sensing imaging technology,
high-resolution images can be easily obtained, which help us understand the earth better.
Object detection, change detection, semantic segmentation and other tasks become popular in remote sensing area.

\cite{han2015object, long2017accurate, bai2014vhr, zhang2016weakly, lei2012rotation} propose different approaches for object detection in remote sensing images with the powerful feature extraction capability of deep convolutional neural networks.
However, those methods mainly focus on small region segments comparing to the original large inputs, e.g., usually over than 20 000 $\times$ 20 000 pixels.
Therefore, they cannot scale up to handle such huge images gracefully.
Zhang~\etal~\cite{zhang2016weakly} attempted to detect airports at first to speed-up the process in large-scale images,
but the training and testing images are all from regions near to airports, which escape from the complex backgrounds.
According to our experiments, it is not robust enough for practical applications.

Object detection in large-scale remote sensing images is pretty challenging.
First, the scale of the input image is too large to reach the practical application.
The computation time and memory consumption are increased quadratically, making it too slow and not runnable on current hardware.
Second, massive and complex backgrounds that appear in a real scenario may introduce more false positives, for instance, desert region with random texture or urban area with massive building structure.
Moreover, the performance drops drastically with tiny objects (such as 8-32 pixels), especially in low-resolution images, which further increases the difficulty of tiny object detection in remote sensing images.

\begin{figure}[t]
	\begin{center}
		\includegraphics[width=1.0\linewidth]{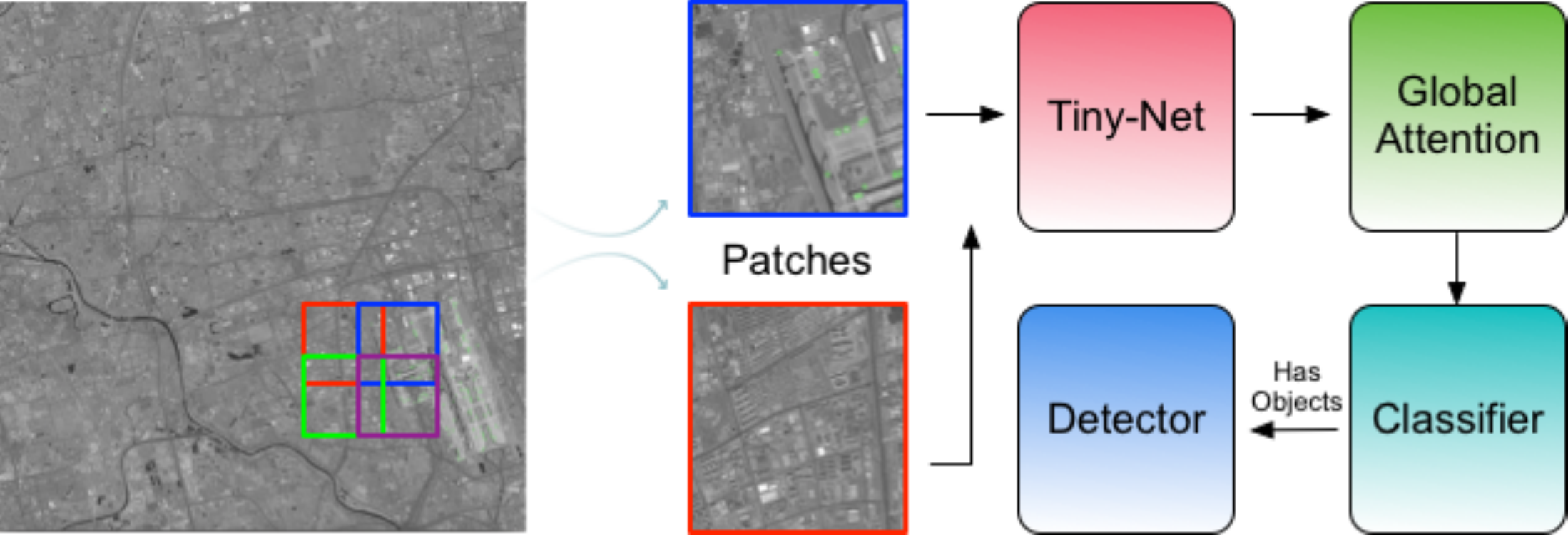}
	\end{center}
	\caption{$\mathcal{R}^2$-CNN is a unified network working in a patch-wise manner. Pipeline is shown on the right. }
	\label{fig.pipeline}
\end{figure}

To tackle these problems, we propose
a unified and self-reinforced convolutional neural network called $\mathcal{R}^2$-CNN: Remote sensing Region-based Convolutional Neural Network,
which is composed of backbone Tiny-Net, intermediate global attention block, and the final classifier and detector,
enable the entire network \emph{efficient} in both computation and memory consumption, \emph{robust} to false positives and \emph{strong} to detect tiny objects.
Pipeline is shown in Fig~\ref{fig.pipeline}.

Firstly, as a unified and self-reinforced framework, $\mathcal{R}^2$-CNN first crops large scale images with a much more smaller scale (such as 640 $\times$ 640 pixels) with 20$\%$ overlap to tackle the oversized input size.
By processing the patches asynchronously, the limited memory is not a problem anymore.
A convolutional backbone structure is then applied to inputs, which enables powerful features extraction.
Based on the discriminative features, a classifier first predicts the existence of detection target in the current patch, and a detector is followed to locate them accurately if available.
The classifier and detector are mutually reinforced each other under the end-to-end training framework. There are two advantages of this self-reinforced architecture as follows:
\vspace{5pt}
\begin{enumerate}
	\item Since, in large-scale remote sensing images, most crops do not contain valid target so that about 99$\%$ of the total patches do not need to pass the heavy detector branch. The light classifier branch can filter out a blank patch without heavier detector cost.
	\item As most false positives commonly occur with massive backgrounds, benefited from the self-reinforced framework, the classifier can identify the difficult situation even when there is only one tiny object in the patch given the fine-grained features from detector. On the other hand, the detector receives less false positive candidates since most of them are filtered out by the classifier. Even if the patches are distinguished incorrectly by a classifier, the detector can still rectify the results later.
\end{enumerate}

Second, we specially designed a lightweight residual net- work called Tiny-Net to reduce the inference cost and preserve powerful features for object detection.
Tiny-Net is motivated by \cite{he2016deep} but is much more lightweight.
On the other hand, Tiny-Net can be trained from scratch with a cycle training schedule because of fewer parameters,
making that the framework does not be influenced by the limited training samplers and the domain gap between natural images and remote sensing images.

Third, to further inhibit the false positives, we also use feature pyramid pooling as a global attention block on the top of Tiny-Net.
The feature maps are first pooled in different pyramid levels, such as $1 \times 1$, $2 \times 2$, and $4 \times 4$.
Then, we recover the pooled features to their original scale with bilinear interpolation.
The feature maps are fused additionally next.
Feature maps get more context information, and the receptive field is also enlarged to the whole image.
The detector is more discriminative with the help of more context information.
We can find that the confidence of false positives drop obviously with this module, proving its effectiveness.

Finally, to make the framework strong to detect tiny objects, we comprehensively analyzed why the detected performance drops drastically with tiny objects and proposed a scale-invariant anchor strategy to tile anchors reasonably, especially for small objects based on the region proposal network (RPN) in \cite{ren2015faster}.
On the other hand, we insert an efficient zoom-out and zoom-in architecture in Tiny-Net to enlarge the feature maps, which improve the recall of tiny objects obviously.
Position-sensitive region of interests (RoI) pooling \cite{dai2016rfcn} is also used to share the computation from all detectors on the entire image and get more spatial information, which is faster and more accurate than the original RoI pooling in \cite{ren2015faster}.

Our contributions can be summarized into four components:
\begin{enumerate}
	\item We proposed a unified and self-reinforced framework called $\mathcal{R}^2$-CNN, which is \emph{efficient} in computation and memory consumption, \emph{robust} to false positives and \emph{strong} to detect tiny objects.
	\item We proposed Tiny-Net, a lightweight residual network which can be trained from scratch and further improve the efficiency.
	\item We insert a global attention block into $\mathcal{R}^2$-CNN to further inhibit the false positives.
	\item We comprehensively analyze why the detected performance drops drastically with tiny objects and then make the framework strong to detect tiny objects.
\end{enumerate}

The remainder of this paper is organized as follows. In section  \uppercase\expandafter{\romannumeral2}, we briefly introduce state-of-art object detection methods and their applications on remote sensing systems.
Then we explain the details of our $\mathcal{R}^2$-CNN in section \uppercase\expandafter{\romannumeral3},
and show the experiments in section \uppercase\expandafter{\romannumeral4}.
Finally, section \uppercase\expandafter{\romannumeral5} concludes this paper with a discussion of the results.

\begin{figure*}[t]
	\begin{center}
		\includegraphics[width=1.0\linewidth]{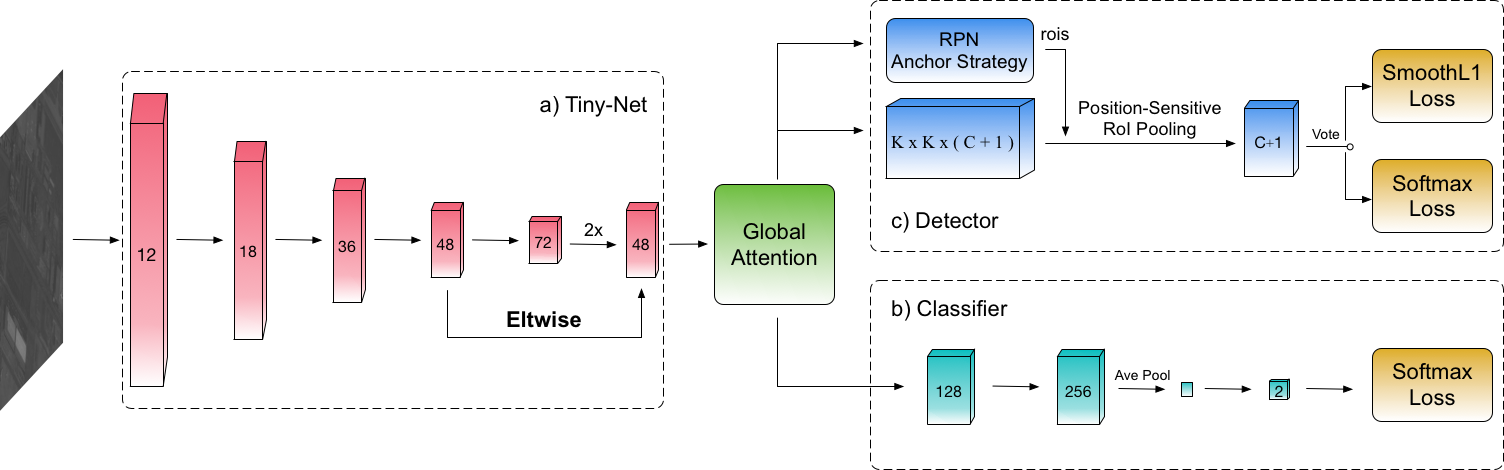}
	\end{center}
	\caption{Architecture of our $\mathcal{R}^2$-CNN. (a) Tiny-Net, a lightweight residual structure which enables fast and powerful features extraction from inputs. (b) Classifier, which can speed-up the unified network and avoid false alarm raised by massive backgrounds. (c) Detector, which can locate target objects accurately if available. The classifier and detector are mutually reinforced each other under the end-to-end training framework. In addition, global attention block is built on top of Tiny-Net to inhibit false positives.}
	\label{fig.architecture}
\end{figure*}

%% file: sections/related_work.tex

\section{Related Work}

As a fundamental problem in a remote sensing area,
object detection in remote sensing images has been extensively studied in recent years.
Previous methods (such as scale-invariant feature transform~\cite{lowe2004sift} and histogram of oriented gradient (HoG)~\cite{hog},~\cite{xiao2015elliptic}) use low-level or middle-level feature representations to detect objects.
Recently, impressive improvements have achieved with convolutional neural networks.
Cheng and Han~\cite{cheng2016survey} provide a review of the recent progress in those fields and propose two promising research directions,
which are deep learning-based methods and weakly supervised learning-based methods.

Convolutional neural networks got a start from LeNet~\cite{lecun1998gradient} and became popular with AlexNet~\cite{krizhevsky2012imagenet}.
Many impressive methods are proposed to promote the development of image recognition from then on, such as network-in-network~\cite{lin2013network}, VGGNet~\cite{simonyan2014vgg}, and GoogLeNet~\cite{szegedy2015going}. ResNet~\cite{he2016deep} is a milestone, which is using residual connections to train very deep convolutional models. It made a great improvement in image recognition.
Object detectors, such as OverFeat~\cite{sermanet2013overfeat} and region convolutional neural network (R-CNN)~\cite{girshick2014rich}, made dramatic improvements in accuracy with those deep learning- based feature representations.
OverFeat adopted a Conv-Net as a sliding window detector on an image pyramid.
R-CNN adopted a region proposal-based method based on selective search~\cite{uijlings2013selective} and then used a Conv-Net to classify the scale-normalized proposals.
Spatial pyramid pooling (SPP)~\cite{he2014spatial} adopted R-CNN on feature maps extracted on a single image scale, which demonstrated that such region-based detectors could be applied much more efficiently.
Fast R-CNN~\cite{girshick2015fast} and Faster R-CNN~\cite{ren2015faster} made a unified object detector in a multitask manner. Region proposal networks are proposed to replace selective search.
Dai~\etal~\cite{dai2016rfcn} proposed R-FCN, which uses position-sensitive RoI pooling to get a faster and better detector.
While those region-based methods are too slow for practical use, a single-stage detector, such as YOLO~\cite{yolo} and SSD~\cite{ssd}, is proposed to accelerate the processing speed but with a performance drop, especially in small objects.

Along with the rapid development with those mechanisms,
small object detection seems much more difficult, and thus, researchers proposed many frameworks for small object detection specifically.
Those methods mainly focus on how to implement a multiscale framework elegantly or using hard mining method which let the network pay more attention to small objects.
Lin~\etal~\cite{fpn} proposed feature pyramid networks that use the top-down architecture with lateral connections as an elegant multiscale feature warping method.
Zhang~\etal~\cite{s3fd} proposed which proposed a scale-equitable face detection framework to handle different scales of faces well.
Hu and Ramanan~\cite{findingtiny} showed that the context is crucial and defines the templates that make use of massively large receptive fields.
Zhao~\etal~\cite{pspnet} proposed a pyramid scene parsing network that employs the context reasonable.
Shrivastava~\etal~\cite{ohem} proposed an online hard example mining method that can improve the performance of small objects obviously.

Many methods~\cite{han2015object, hu2015transferring, zhang2015hierarchical, ishii2015surface, vsevo2016convolutional, salberg2015detection, zhang2016weakly, cheng2016learning, long2017accurate, lei2012rotation, bai2014vhr}
are proposed to improve the object detection accuracy in remote sensing images with convolutional neural networks.
Those methods often use pretrained CNN models on large data sets to handle the limited remote sensing training data.
Zhang~\etal~\cite{zhang2015hierarchical} used the trained CNN models to extract surrounding features. Those features were combined with features from HoG to get final representations and then applied gradient orientation to generate region proposals.
Zhu~\etal \cite{zhu2015orientation} used CNN features from multilevel layers for object detection, which handle the scale-invariance with single scale input.
Jiang~\etal \cite{jiang2015deep} used a graph-based superpixel segmentation to generate proposals and then trained a CNN to classify these proposals into different classes.
Cheng~\etal \cite{cheng2016learning} introduced a rotation-invariant operator to the existing CNN architectures and achieves a significant performance.
Long~\etal \cite{long2017accurate} proposed an unsupervised score- based bounding-box regression for accurate object localization in remote sensing images.
Those methods mainly focus on small region segment compared to the original large remote sensing image input, usually over 10 000 $\times$ 10 000 pixels, and thus, they cannot scale up to handle such large input gracefully.
Zhang~\etal \cite{zhang2016weakly} attempted to detect airports in large-scale images first to reduce overall airplane detection time, but the training and testing images are all from the region near airports without arbitrary massive backgrounds.
For practical use, object detection in large-scale remote sensing images is very important and necessary.

%% file: sections/methodology.tex

\begin{figure*}[t]
\begin{adjustwidth}{1.0cm}{1.0cm}
  \begin{subfigure}[h]{.4\textwidth}
    \centering
    \includegraphics[width=1.0\linewidth]{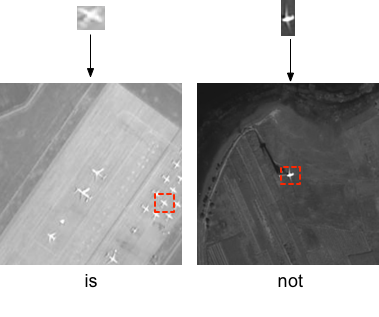}
    \caption{}
  \end{subfigure}
  \hfill
  \begin{subfigure}[h]{.4\textwidth}
    \centering
    \includegraphics[width=0.7\linewidth]{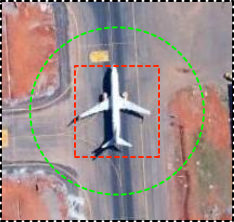}
    \caption{}
  \end{subfigure}

  \medskip

  \begin{subfigure}[h]{.4\textwidth}
    \centering
    \includegraphics[width=1.0\linewidth]{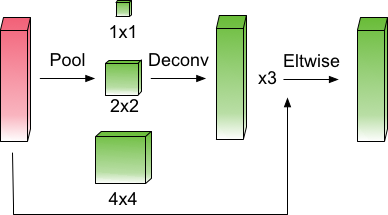}
    \caption{}
  \end{subfigure}
  \hfill
  \begin{subfigure}[h]{.4\textwidth}
    \centering
    \includegraphics[width=1.0\linewidth]{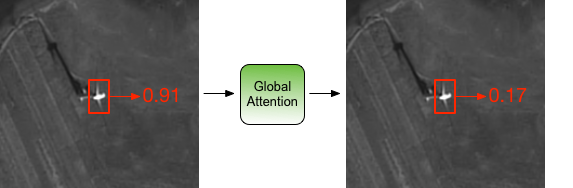}
    \caption{}
  \end{subfigure}

  \caption{(a) Whether they are airplanes or not?
  			(b) Black area: theoretical receptive field. Green area: effective receptive field. Red area: bounding box.
  			(c) Global attention block.
  			(d) False positive's confidence drop obviously with global attention.
  }
  \label{fig.3}
\end{adjustwidth}
\end{figure*}

\begin{figure*}[t]
\begin{adjustwidth}{1.0cm}{1.0cm}
  \begin{subfigure}[h]{.4\textwidth}
    \centering
    \includegraphics[width=1.0\linewidth]{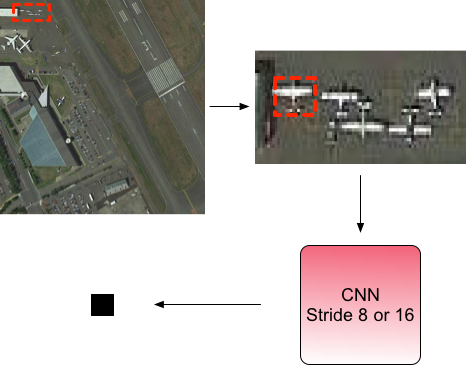}
    \caption{}
  \end{subfigure}
  \hfill
  \begin{subfigure}[h]{.4\textwidth}
    \centering
    \includegraphics[width=1.0\linewidth]{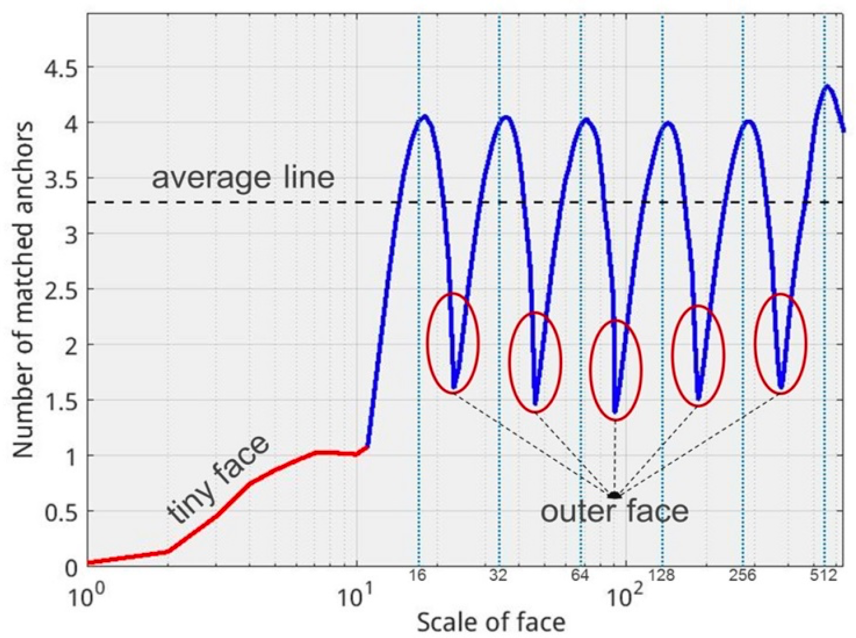}
    \caption{}
  \end{subfigure}

  \medskip

  \begin{subfigure}[h]{.4\textwidth}
    \centering
    \includegraphics[width=1.0\linewidth]{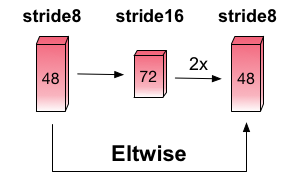}
    \caption{}
  \end{subfigure}
  \hfill
  \begin{subfigure}[h]{.4\textwidth}
    \centering
    \includegraphics[width=1.0\linewidth]{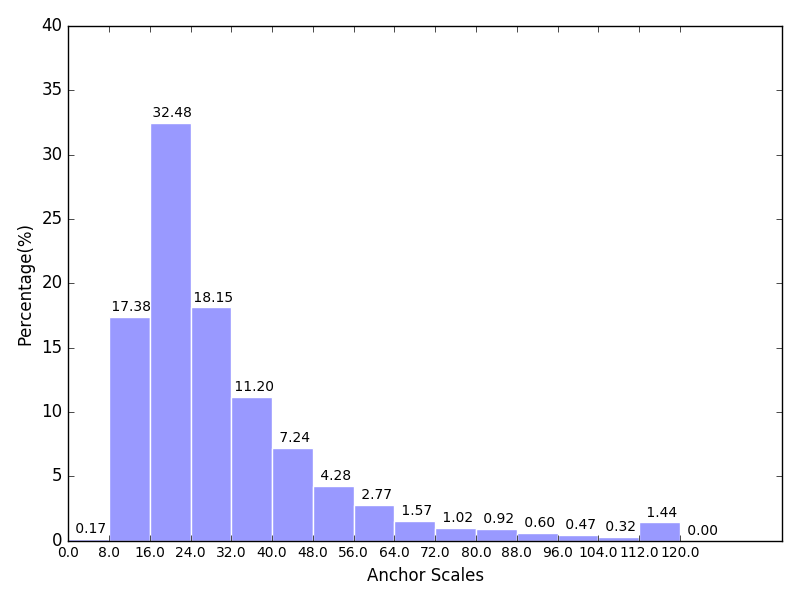}
    \caption{}
  \end{subfigure}

  \caption{(a) Few Features: small objects have a few features at detection layer.
  			(b) Anchor matching analysis: the figure is from $S^3$FD~\cite{s3fd}, and tiny and outer objects match too little anchors.
  			(c) Our skip-connection architecture for reducing the anchor stride by enlarging the feature map.
  			(d) Data Distribution: bounding-box scales of our training data set.
  }
  \label{fig.4}
\end{adjustwidth}
\end{figure*}

\section{Proposed METHOD}

The $\mathcal{R}^2$-CNN, shown in Fig~\ref{fig.architecture}, consists of the backbone Tiny-Net, intermediate global attention block, and final classifier and detector.

\subsection{\bf \emph{$\mathcal{R}^2$-CNN}}
$\mathcal{R}^2$-CNN is a unified and self-reinforced framework working in an end-to-end manner.
Considering that the large input image size increases the computation time and memory consumption quadratically,
large-scale remote sensing images (such as 20 000 $\times$ 20 000 pixels) are cropped with a much more smaller scale (such as 640 $\times$ 640 pixels) with 20$\%$ overlap.
By processing the patches asynchronously, the limited memory is not a problem anymore.

A convolutional backbone structure is then applied to the inputs, which enables powerful features extraction.
Based on those discriminative features, the classifier first predicts the existence of detection target in the current patch, and the detector is followed to locate them accurately is available.
The classifier and detector are mutually reinforced each other under the end-to-end training framework.
There are two advantages of this self-reinforced architecture as follows:

First, the light classifier branch can filter out a blank patch without heavier detector cost.
Classifier's architecture is in Fig~\ref{fig.architecture}-(b), and we just use two CONV-BN-RELU blocks to extract features from the former features.
Global average pooling and a $1\times 1$ convolutional operator are then attached to it.
Softmax loss is employed to guide the training of the classifier.
Considering that most crops do not contain a valid target in remote sensing images, about $99\%$ of the total patches do not need to pass the heavy detector branch.

Second, massive and complex backgrounds appear in a real scenario may introduce more false positive, for instance, desert region with random texture or urban area with massive building structure.
The false positives are first inhibited by the mutual reinforcement from the classifier and the detector.
On the one hand, the classifier can distinguish the difficult situation even when there is only one tiny objects (such as $12 \times 12$ pixels) in the patch.
We explain this promotion mainly given the fine-grained feature extracted from detector.
On the other hand, the detector receives less false positive candidates since most of them are filtered out by classifier.
Even if the patches are distinguished incorrectly by the classifier, the detector can still rectify the results later.

There are three outputs from our network.
One output $m$ from classifier represents the probability of whether there are target objects in corresponding patch or not.
Two outputs from detector represent the discrete probability ($p = (p_0, ..., p_k)$) distribution of each RoI over $K+1$ categories, and bounding-box regression offsets, $t^k = (t^k_x, t^k_y, t^k_w, t^k_h)$, for each of the $K$ object classes, index by $k$, in corresponding patch.
We use the parameterization for $t^k$ in \cite{girshick2014rich}, in which $t^k$ represents a scale-invariant translation and log-space height/width shift relative to an object proposal.
Each training patches is labeled with a binary ground-truth $n$, and each RoI in detector is labeled with a ground-truth class $u$ and a ground-truth bounding-box regression target $v$.
We use a unified multi-task loss $L$ on each patch to joint classifier and detector:
\begin{equation}
	\begin{aligned}
		L(m,n,p,& u,t^u,v) = L_{cls}(m, n) + \\
		& \mu [n = 1] ( L_{cls}(p, u) + \lambda[u \ge 1] L_{loc}(t^u, v) ),
	\end{aligned}
\end{equation}
in which $L_{cls}(p,u)$ and $L_{cls}(m,n)$ are softmax loss and $L_{loc}$ is smooth-L1 Loss in \cite{ren2015faster}. The hyper-parameter $\lambda$ and $\mu$ in (1) controls the balance between the three task losses. All experiment use $\lambda = 1$ and $\mu = 1$.
We only back-propagate detector's loss when there are detection targets in the corresponding patch during training time.
The entire network is \emph{efficient, robust}, and \emph{strong}.

\subsection{\bf \emph{Tiny-Net}}
Recently, CNN-based methods often use VGG~\cite{simonyan2014vgg} or ResNets~\cite{he2016deep} as feature extractors.
Those models are pretrained on ImageNet~\cite{imagenet}, a large-scale hierarchical image database with millions of images, to deal with the limited training samples and get a much more quicker convergence.
However, there are still many disadvantages when using those pre-trained models.
First, those models are too heavy to reach real-time efficiency.
Second, those models are designed specifically for image classification, making that the feature resolution may be not enough for object detection.
Finally, considering the heavy parameters, training scratch is pretty difficult, especially with limited training samples.
When applying the pretrained models to remote sensing frameworks, the domain gap between natural images and remote sensing images may make the models sub-optimal.

The architecture of Tiny-Net is shown in Table~\ref{tab:table1}.
The $3 \times 3$ block is a residual block in ResNet\cite{he2016deep} except conv-1.
We do not apply the downsample operator in conv-1 to enable the feature maps more discriminative for tiny object detection, which is different from ImageNet pretrain models such as VGG~\cite{simonyan2014vgg} and ResNets~\cite{he2016deep}.
The parameters of Tiny-Net are far less than ResNets.
Thanks to this lightweight architecture, Tiny-Net can be trained from scratch and converge well just with a cycle training schedule, which iteratively updates the step learning rate twice or more.
Under this condition, Tiny-Net will not be influenced by the domain gap between the natural images and the remote sensing images.

Benefit from those characters,
Tiny-Net can reduce inference cost and preserve powerful features for tiny object detection in remote sensing images, which further improved the efficiency of $\mathcal{R}^2$-CNN.

\begin{table}[t]
  \renewcommand\arraystretch{1.5}
  \begin{center}
    \caption{Architecture of our Tiny-Net. Each $3 \times 3$ block are residual block in ResNet\cite{he2016deep} except conv-1.}
    \label{tab:table1}
    \begin{tabular}{c|c|c}
      \hline
      \textbf{layer name} & \textbf{output size} & \textbf{architecture}\\
      \hline
      conv-1 & $640 \times 640$ & $[3 \times 3, 12] \times 2$\\
      \hline
      conv-2 & $320 \times 320$ & $\left[ \begin{array}{ccc}
      	                                3 \times 3, 18 \\
      	                                3 \times 3, 18
                                      \end{array} \right]$ $\times$ 2\\
                                      \hline
      conv-3 & $160 \times 160$ & $\left[ \begin{array}{ccc}
      	                                3 \times 3, 36 \\
      	                                3 \times 3, 36
                                      \end{array} \right]$ $\times$ 2\\
                                      \hline
      conv-4 & $80 \times 80$   & $\left[ \begin{array}{ccc}
      	                                3 \times 3, 48 \\
      	                                3 \times 3, 48
                                      \end{array} \right]$ $\times$ 3\\
                                      \hline
      conv-5 & $40 \times 40$   & $\left[ \begin{array}{ccc}
      	                                3 \times 3, 72 \\
      	                                3 \times 3, 72
                                      \end{array} \right]$ $\times$ 2\\
      \hline
    \end{tabular}
  \end{center}
\end{table}

\subsection{\bf \emph{Global Attention}}
Thanks to the unified classifier and detector, numerous blank patches are filtered by classifier, making the false positives reduce obviously.
However, the problem still exists because of the limited receptive field.
When you catch sight of two objects with similar appearance, you may not be sure what they are without context information.
For example, when you see the top image in Fig~\ref{fig.3}-(a), you may confuse in this question: what exactly are they?
However, when you see two images in the bottom image, you can easily distinguish them out.
As discussed in \cite{receptive}, the CNN has two types of receptive fields: the theoretical receptive field and the effective receptive field.
The Theoretical receptive field indicates the input region that can affect the value of this unit theoretically.
However, not every pixel in the theoretical receptive field contributes equally to the final output.
Only a subset of the area has an effective influence on the output value, which is called effective receptive field.
The effective receptive field is smaller than the theoretical receptive field, as shown in Fig~\ref{fig.3}-(b).
The limited effective receptive field leads the final feature map to obtain little context information, thus leading to more false positives.

Inspired by this phenomenon, we use feature pyramid pooling as a global attention block on the top of Tiny-Net.
The architecture is shown in Fig~\ref{fig.3}-(c).
The feature maps are first pooled in different pyramid levels, such as $1\times1, 2\times2$, and $4\times4$.
Then, we recover the pooled features to their original scale with bilinear interpolation.
The feature maps are fused additionally next. Feature maps can get more context information, and the receptive field will also be enlarged to the whole image.
The global attention module fuses the features from different pyramid scales and leads the detector to pay more attention to the whole image.
The detector is more discriminative with the help of more context information.
We can found that the confidence of false positives drops obviously with this module, as shown in Fig~\ref{fig.3}-(d), proving its effectiveness.

\begin{table*}[t]
  \renewcommand\arraystretch{1.5}
  \begin{center}
    \caption{$\mathcal{R}^2$-CNN's results in \emph{GF1-test-dev}, results are shown in unified / un-unified / fully detection training and testing.}
    \label{tab:r2cnn-1}
    \begin{tabular}{c|c|c|c|c}
      \hline
      \textbf{Score Thre} & \textbf{TP} & \textbf{FP}& \textbf{Recall}& \textbf{Precision}\\
      \hline
  0.05 & 613 / 593 / 616 & 186 / 264 / 8126 & 99.35 / 96.11 / 99.84 & 76.72 / 69.19 / 6.97 \\
  0.5  & 607 / 591 / 612 & 46 / 93 / 561  & 98.38 / 95.79 / 99.19 & 92.96 / 86.40 / 52.17 \\
  0.8  & 593 / 577 / 596 & 15 / 41 / 264    & 96.11 / 93.52 / 96.60 & 97.53 / 93.36 / 69.30 \\
  0.85 & 591 / 573 / 592 & 10 / 29 / 189    & 95.78 / 92.87 / 95.95 & 98.33 / 95.18 / 75.80 \\
  0.9  & 579 / 568 / 586 & 7 / 18 / 131     & 93.84 / 92.06 / 94.98 & 98.80 / 96.93 / 81.73 \\
  0.95 & 556 / 542 / 561 & 3 / 7 / 37       & 90.11 / 87.84 / 90.92 & 99.46 / 98730 / 93.81 \\
      \hline
    \end{tabular}
  \end{center}
\end{table*}

\begin{table*}[t]
  \renewcommand\arraystretch{1.5}
  \begin{center}
    \caption{$\mathcal{R}^2$-CNN's results in \emph{GF2-test-dev}, results are shown in unified / un-unified / fully detection training and testing.}
    \label{tab:r2cnn-2}
    \begin{tabular}{c|c|c|c|c}
      \hline
      \textbf{Score Thre} & \textbf{TP} & \textbf{FP}& \textbf{Recall}& \textbf{Precision}\\
      \hline
  0.05 & 108 / 79 / 113 & 74 / 173 / 5746 & 93.91 /  68.69  / 98.26 & 59.34 / 31.35 / 1.93 \\
  0.5  & 107 / 77 / 111 & 14 / 34 / 217   & 93.04 /   66.95  / 96.52 & 88.43 / 69.36 / 33.84 \\
  0.8  & 105 / 74 / 108 & 8 / 21 / 69     & 91.30 /  64.34 / 93.91 & 92.92 / 77.89 / 61.01 \\
  0.85 & 104 / 74 / 106 & 5 / 17 / 47     & 90.43 /  64.34  / 92.17 & 95.41 / 81.32 / 69.28 \\
  0.9  & 100 / 72 / 105 & 2 / 13 / 31     & 86.96 /  62.60  / 91.30 & 98.04 / 84.71 / 77.20 \\
  0.95 & 96 / 69 / 100 & 1 / 7 / 13        & 83.48 / 60.00 / 86.96 & 98.97 / 90.79 / 88.49 \\
      \hline
    \end{tabular}
  \end{center}
\end{table*}

\subsection{\bf \emph{Detector}}
Our $\mathcal{R}^2$-CNN is strong to detect tiny objects.
The architecture of detection branch is shown in Fig~\ref{fig.architecture}-(c).
The state-of-the-art object detectors are mainly based on the RPN in \cite{ren2015faster} framework, which use anchors to generate object proposals.
Anchors are a set of predefined boxes with multiple scales and aspect ratios tiled regularly on the image plane.
However, anchor-based detectors drop the performance drastically on objects with tiny sizes, such as less than $16 \times 16$ pixels, and those tiny objects are majority in remote sensing images, such as airplanes, ships, and cars.
To tackle this problem, we first investigate why this is the case
and proposed a scale-invariant anchor strategy to tile anchors reasonably, especially for small objects.
On the other hand, we insert an efficient zoom-out and zoom-in architecture in Tiny-Net to enlarge the feature map without margin cost, which improves the recall of tiny objects obviously.
Position sensitive RoI pooling is also used to get more spatial information.
Through these ways, we get the excellent results on tiny object detection.

\subsubsection{\emph{Why this is the case?}}
Like in Fig~\ref{fig.4}-(a), the stride size of the lowest anchor-associated layer is too large (e.g., 16 pixels or 32 pixels),
and the features loss along with the downsampling of pooling layer.
Therefore, tiny objects have been highly squeezed on these layers and have a few features for detection.
An airplane may be only $1 \times 1$ pixels in final feature map.
On the other hand, the anchor scales are discrete (i.e., $ 16, 32, 64, ... 2^k)$, but object scales are continuous.
During training, an anchor will be assigned to a ground-truth box if its intersection over union (IoU) with this box is the highest than other anchors,
or its IoU is higher than a threshold $T_h$.
When the object's scale is near to anchor scales, they will be attached more anchors, thus easier to be located.
The face detector $S^3$FD～\cite{s3fd}, which uses SSD～\cite{ssd} architecture explained this phenomenon appropriately, and we infer their statistics in Fig~\ref{fig.4}-(b).
It shows the number of matched anchors at different face scales under $ 16, 32, 64, ... 2^k $ anchor scales.
If an object has a scale over the average line, it will be matched enough anchors.
However, tiny objects are matched a few anchors, leading to performance drop drastically.

\subsubsection{\emph{Our Method}}

To tile anchors more reasonably, we analyze the bounding-boxes' scale distribution of our training data set, which is shown in Fig~\ref{fig.4}-(d).
We can see that tiny objects are majority in it.
Instead of choosing anchors by hand, we run k-means clustering on the training set to automatically find good priors.
Our training set's bounding box scales are $X: (x1, x2, ..., x_n)$.
There are $k$ anchor scales we want, and the center scales are $(\mu_1, \mu_2, ..., \mu_k)$.
Scales are clustered in $(s1, s2, ..., s_k)$.
Through minimizing:
\begin{equation}
	\centering
	arg min{\sum_{i=1}^{k}{\sum_{x\in s_i}{||x-\mu_i||^2}}},
\end{equation}
we can get the clustered anchor scales.

To extract beneficial features for tiny object detection,
one way is to reduce the anchor stride by enlarging the feature map using a zoom-out and zoom-in architecture.
Like shown in Fig~\ref{fig.4}-(c), we first zoom out the feature map with a residual block, and thus, its anchor-stride is 16 pixels.
Then, we employed to recover the feature maps that are recovered to their original scale using a zoom-in operator.
In addition, we use a skip connection between the stride-8 layer and the upsampled layer.
We found that the stride-16 layer can extract more high-level features, which is beneficial to object detection.
The skip-connection can fuse low-level features and high-level features, making final feature maps more discriminative.

Considering the complicated backgrounds of remote sensing images,
we use position-sensitive RoI pooling\cite{dai2016rfcn} in our detector instead of RoI pooling.
RoI pooling applies costly per-region subnetwork hundreds of times.
If there are 1000 proposals, the detector will be tested 1000 times wastefully.
In contrast to this operator, position-sensitive RoI pooling is fully convolutional with almost all computation shared on the entire image.
Much computation is saved with this operator.
On the other hand, position sensitive RoI pooling can address the dilemma between translation-invariance in image classification and translation-variance in object detection.
More spatial information is extracted, thus leading to better performance.

%% file: sections/experiments.tex
\section{Experiments}
In this section, we first present the implementations of our $\mathcal{R}^2$-CNN, such as datasets, evaluation metric and parameter settings. The results of our network and comparative experiments are then discussed.

\subsection{\bf \emph{Implementations}}
\subsubsection{\bf \emph{Datasets}}
Due to the lack of standard data sets of large-scale remote sensing images, we collected 1169 GF-1 images and 318 GF-2 images that are 18 000 $\times$ 18 192 pixels, 2.0-m resolution and 27 620 $\times$ 29 200 pixels, 0.8-m resolution, respectively.
In addition, we collected 38 472 pieces of 640 $\times$ 640 images, which contains target objects from the publicly available Google Earth service to supplement the poor positive patches in GF images.
All images are cropped in the patches of 640 $\times$ 640 pixels with $20\%$ overlap.
In particular, if we have a maximum object scale d and a cropped scale D, we recommend an overlap $\frac{d}{D}$.
For example, as the scales in Fig.4-(d), the maximum scale of the objects in our data set is about 128 and the cropped scale is 640, thus, a 20$\%$ overlap is applied to not only prevent the object from being truncated off but also augment the data sets.
To help the convergence of the network, we control the proportion of positive patches and negative patches to be $1:3$ to obtain a balanced training set.
The negative patches are selected using hard example boosting to enhance the training process.

We collect 102 GF-1 images as \emph{GF1-test-dev} and 40 GF-2 images as \emph{GF2-test-dev} to evaluate the ability of $\mathcal{R}^2$-CNN.
To evaluate our model more exhaustively, we collected 4633 images ($640 \times 640$ pixels) from Google Earth as \emph{Rgb-test-dev} and 1000 images ($640 \times 640$ pixels) from GF-1 and GF-2 as \emph{Gray-test-dev}, which help us evaluate the ability of the detector.

\subsubsection{\bf \emph{Evaluation Metric}}
Considering of the few target objects in large-scale remote sensing images and the requirement for practical engineering applications,
we evaluation our $\mathcal{R}^2$-CNN with \emph{Recall} and \emph{Precision} of different score thresholds.
The correct number of detections is true positives $TP$, and the number of spurious detections of the same object is false positives $FP$.
The number of ground-truth instances is $NP$.
The precision and recall following:
\begin{equation}
	Precision = \frac{TP}{TP + FP},
\end{equation}
\begin{equation}
	Recall = \frac{TP}{NP},
\end{equation}
We also show the instance number in detail for more intuitional and convincing.
Considering numerous negative scene of classifier, the \emph{Accuracy} are generally over 99.0$\%$, making this metric meaningless.
Therefore, we use \emph{Recall}, \emph{Precision} and instance number to evaluate our classifier.
We use \emph{mAP} and \emph{Max-Recall} with a score threshold of 0.05 as in PASCAL-VOC \cite{voc} to evaluate our detector, which are defined as:
\begin{equation}
	mAP = \frac{1}{Q} \sum^{Q}_{q=1}AP(q),
\end{equation}
\begin{equation}
	AP = \frac{\sum^n_{k=1}(P(k) \times r(k))}{\mid R(q) \mid},
\end{equation}
where $Q$ is the number of categories, $\mid R(q) \mid$ is the number of images relevant to the category, $q$, $k$ is the rank in the sequence of retrieved images, and $n$ is the total number of retrieved images. $P(k)$ is the precision at cutoff $k$ in the list, and $r(k)$ is an indicator function whose value is 1 if the image at rank $k$ is relevant and is 0 otherwise.
\subsubsection{\bf \emph{Parameter Settings}}
We adopt synchronized stochastic gradient descent training on 8 GPUs with synchronized batch normalization. A mini-batch involves 1 image per GPU and 512 proposals per image for detector training.
We use a momentum of 0.9 and a weight decay of 0.0005. We use a learning rate of 0.001 for 80k minibatches, and 0.0001, 0.00001 for the next 80k and 40k mini-batches.
The learning rate and training epochs are iterated twice as a cycle schedule, because the network is trained from scratch.
We randomly initialize all layers by drawing weights from a zero-mean Gaussian distribution with a standard deviation of 0.01.
The anchor's scales in RPN we used in our final model is $[1, 2, 4, 10, 16, 30]$ with a stride of 8, and the anchor's ratio is 1 considering of airplane's shape.
Other implementation details are as the same as \cite{he2016deep} and \cite{ren2015faster}.
We also use multi-scale training with scale $[513, 609, 721, 801, 913]$,
and online hard example mining \cite{ohem} for hard example boosting.
We stretch remote sensing images from uint16 to uint8 which is divided by a factor of 4.
We also stretch the images' histogram with a factor in $[0, 0.02]$,
which can not only inhibit the noise in remote sensing images but also as a data augmentation method.
We augment the data online with rotation and flipping randomly. Our implementation use Caffe\cite{caffe}.

\begin{table}[t]
  \renewcommand\arraystretch{1.5}
  \begin{center}
    \caption{Results of classifier on \emph{GF1-test-dev}. $\vdash$ means finetuning from ImageNet pretrained model. $\dashv$ means training from scratch. $-$ means training without detector.}
    \label{tab:cls-1}
    \begin{tabular}{c|c|c|c|c|c}
      \hline
      \textbf{Model} & \textbf{TP} & \textbf{FP} & \textbf{Recall}& \textbf{Precision} & \textbf{Time Cost} \\
      \hline
      	ResNet-50$^{\vdash-}$\cite{he2016deep} &  134 & 34  & 87.01 & 79.76 & 48.21 ms\\
      	ResNet-50$^{\dashv-}$\cite{he2016deep} &  128 & 29  & 83.12 & 81.53 & 48.21 ms\\
      	Ours$^-$                            &  105 & 179 & 68.18 & 36.97 & 16.63 ms\\
      \hline
      	$\mathcal{R}^2$-CNN                 & 148 & 45 & 96.10 & 76.68   & 16.63 ms\\
      \hline
    \end{tabular}
  \end{center}
\end{table}

\begin{table}[t]
  \renewcommand\arraystretch{1.5}
  \begin{center}
    \caption{Results of classifier on \emph{GF2-test-dev}. $\vdash$ means finetuning from ImageNet pretrained model. $\dashv$ means training from scratch. $-$ means training without detector.}
    \label{tab:cls-2}
    \begin{tabular}{c|c|c|c|c|c}
      \hline
      \textbf{Model} & \textbf{TP} & \textbf{FP} & \textbf{Recall}& \textbf{Precision} & \textbf{Time Cost} \\
      \hline
      	ResNet-50$^{\vdash-}$\cite{he2016deep} & 61 & 2    & 68.54 & 96.83 & 48.21 ms\\
      	ResNet-50$^{\dashv-}$\cite{he2016deep} & 82 & 204  & 92.13 & 28.67 & 48.21 ms \\
      	Ours$^-$                            & 55 & 67   & 61.80 & 43.65 & 16.63 ms \\
      \hline
      	$\mathcal{R}^2$-CNN                 & 82 & 38   & 92.13 & 68.33 & 16.63 ms\\
      \hline
    \end{tabular}
  \end{center}
\end{table}

\subsection{\bf \emph{Evaluation of $\mathcal{R}^2$-CNN}}
We comprehensively evaluate our method on the \emph{GF1-test-dev} and \emph{GF2-test-dev}.
The results are shown in Table~\ref{tab:r2cnn-1} and Table~\ref{tab:r2cnn-2}.
It is \emph{efficient} that we can process a 18 000 $\times$ 18 192 GF-1 image in 29.4s on Titian X just with single thread.
It is \emph{robust} that with a score threshold of 0.85, there are only 10 false positives in \emph{GF1-test-dev} and 5 false positives in \emph{GF2-test-dev}.
It is \emph{strong} that we can get 95.78 \emph{Recall} and 98.33 \emph{Precision} on \emph{GF1-test-dev} also with a score threshold of 0.85, showing its potential from practical application.
We compared our $\mathcal{R}^2$-CNN with un-unified (train and test separately) network and fully detection network.
The results show when we joint detector with classifier, they can promote each other to get the best results.
When the network is un-unified, the classifier cannot obtain enough discrimination without the feature extracted by detector, leading \emph{Recall} and \emph{Precision} drop obviously.
When detecting the objects all using detector, too many false positives appeared and the efficiency is lower considering of the heavy detector.
Our network achieves the superior results on performance and speed, showing its \emph{efficient, robust}, and \emph{strong}.

\subsubsection{\bf \emph{Efficient}}
Tiny-Net enables the inference time of network less than ResNet-50 or other large models and preserves powerful features for object detection, and the details are shown in Table~\ref{tab:cls-1} and Table~\ref{tab:cls-2}.
The classifier with Tiny-Net costs 16.63 ms with $640 \times 640$ inputs on Titian X, and the detector costs 45.21 ms with the same setting.
The detector is three times slower than classifier.
In our \emph{GF1-test-dev}, there are only 154 patches that have target objects.
The unified architecture makes $99.9\%$ of the total patches do not need to pass the heavy detector branch.
Total costs of our network and fully detection network are shown in Table~\ref{tab:time}.
We can process a 18 000 $\times$ 18 192 GF-1 image in 29.4s on Titian X just with single thread.
Though there is still a long way to build a real-time detection system on large-scale remote sensing images,
our network tackles the problem well with the proposed methods.

\begin{table}[t]
  \renewcommand\arraystretch{1.5}
  \begin{center}
    \caption{Time costs of different methods with single thread.}
    \label{tab:time}
    \begin{tabular}{c|c|c}
      \hline
      \textbf{Benchmark} & \textbf{$\mathcal{R}^2$-CNN} & \textbf{Detection} \\
      \hline
      	\emph{GF1-test-dev} & 29.4 s & 64.7 s \\
      	\emph{GF2-test-dev} & 66.2 s & 163.7 s\\
      \hline
    \end{tabular}
  \end{center}
\end{table}

\begin{table}[t]
  \renewcommand\arraystretch{1.5}
  \begin{center}
    \caption{Performance with or without global attention block.}
    \label{tab:global}
    \begin{tabular}{c|c|c|c|c}
      \hline
      \textbf{Score Thre} & \textbf{TP} & \textbf{FP}& \textbf{Recall}& \textbf{Precision}\\
      \hline
      	0.05 & 613 / 611 & 186 / 216 & 99.35 / 99.03 & 76.72  / 73.88   \\
      	0.5  & 607 / 609 & 46 / 101   & 98.38 / 98.70 & 92.96 / 85.77   \\
      	0.8  & 593 / 600 & 15 / 63   & 96.11 / 97.24 & 97.53  / 90.49   \\
      	0.85 & 591 / 597 & 10 / 49    & 95.78 / 96.76 & 98.33 / 92.41   \\
      	0.9  & 579 / 582 & 7 / 27     & 93.84 / 94.33 & 98.80 /95.56   \\
      	0.95 & 556 / 559 & 3 / 11     & 90.11 / 90.60 & 99.46 / 98.07   \\
      \hline
    \end{tabular}
  \end{center}
\end{table}

\begin{table*}[t]
  \renewcommand\arraystretch{1.5}
  \begin{center}
    \caption{Results of detectors on \emph{Rgb-test-dev}. - means training without classifier. + means training with ImageNet pre-trained model.}
    \label{tab:det-1}
    \begin{tabular}{c|c|c|c|c|c|c|c|c|c}
      \hline
      \textbf{Metric} & \textbf{stride-16} & \textbf{No Conv-5} & \textbf{ResNet-50$^+$}\cite{dai2016rfcn}
      & \textbf{Faster R-CNN}\cite{ren2015faster}
      & \textbf{Anchor-4} & \textbf{Anchor-5} & \textbf{Anchor-8}
      & \textbf{Ours$^-$} & $\mathcal{R}^2$-CNN \\
      \hline
      	mAP ($\%$)  &  84.73 & 93.74 & 83.86 & 94.61 & 94.95 & 95.12 & 96.43 & 95.81  & 96.04 \\
      \hline
      	Max Recall ($\%$)  &  87.21 & 95.40 & 86.34 & 95.62 & 96.38 & 96.74 & 97.88 & 97.12 & 97.26 \\
      \hline
      	Time cost (ms)   &  31.93 & 41.81  & 58.26  & 49.26  & 39.58 & 42.53 &  50.07 & 45.21 & 45.21 \\
      \hline
    \end{tabular}
  \end{center}
\end{table*}

\begin{table*}[t]
  \renewcommand\arraystretch{1.5}
  \begin{center}
    \caption{Results of detectors on \emph{Gray-test-dev}. - means training without classifier. + means training with ImageNet pre-trained model.}
    \label{tab:det-2}
    \begin{tabular}{c|c|c|c|c|c|c|c|c|c}
      \hline
      \textbf{Metric} & \textbf{stride-16} & \textbf{No Conv-5} & \textbf{ResNet-50$^+$}\cite{dai2016rfcn}
      & \textbf{Faster R-CNN}\cite{ren2015faster}
      & \textbf{Anchor-4} & \textbf{Anchor-5} & \textbf{Anchor-8}
      & \textbf{Ours$^-$} & $\mathcal{R}^2$-CNN \\
      \hline
      	mAP ($\%$) &  83.68 & 95.16 & 82.53 & 94.90 & 94.56 & 95.35 & 97.47 & 97.01 & 97.25 \\
      \hline
   Max Recall ($\%$)  &  85.19 & 96.15 & 84.07 & 96.04 & 95.58 & 96.49 & 98.23 & 98.13 & 98.03 \\
      \hline
      	Time cost (ms)   &  31.93 & 41.81 & 58.26 & 49.26 & 39.58 & 42.53 & 50.07 & 45.21 & 45.21\\
      \hline
    \end{tabular}
  \end{center}
\end{table*}

\subsubsection{\bf \emph{Robust}}
The unified classifier can identify the difficult situation even when there is only one tiny object in the patch, given the fine-grained feature extracted by detector,
and the detector receives less false positive candidates since most of them are filtered out by the classifier.
Results are shown in Table~\ref{tab:cls-1} and Table~\ref{tab:cls-2}.
Because of the large memory needed by ResNet-50 with an anchor-stride of 8,
we cannot evaluate our $\mathcal{R}^2$-CNN with ResNet-50.
Thus, the setting of ResNet-50 is the same as in \cite{he2016deep} without detector.
\emph{GF1-test-dev} is cropped in 131 148 patches with $640 \times 640$ pixels and $20\%$ overlap,
and there are also 123 120 patches from \emph{GF2-test-dev}.
The \emph{Recall} and \emph{Precision} only consider the positive patches.
The results of our $\mathcal{R}^2$-CNN get the best results compared with others.
The classifier drop performance drastically without detector,
and the recall of our $\mathcal{R}^2$-CNN is higher than ResNet-50,
showing the effectiveness of the features extracted by detector.
Though there are more false positives in our model,
the detector can rectify the results later.
We have attempted to train ResNet-50 from scratch to break the domain gap between natural images and remote sensing images but get bad results.
We argue that this is mainly because the numerous parameters of ResNet-50 but with the small amount of training sets.
Though there are hundreds of thousands remote sensing images for us,
only a few of them can be used during training time to handle the positive-negative imbalance problem.

Besides, we evaluate the effectiveness of global attention block on \emph{GF1-test-dev}.
The results are shown in Table~\ref{tab:global}.
When $\mathcal{R}^2$-CNN is implemented without global attention block, there are more false positives because of the limited receptive field.
With the global attention block, the confidence of false positives drop obviously.
We found that the recall drops a little with global attention block.
Comparing to the enhancement of inhibiting false positives,
this is a better model for practical engineering.

\begin{figure}[t]
	\begin{center}
		\includegraphics[width=1.0\linewidth]{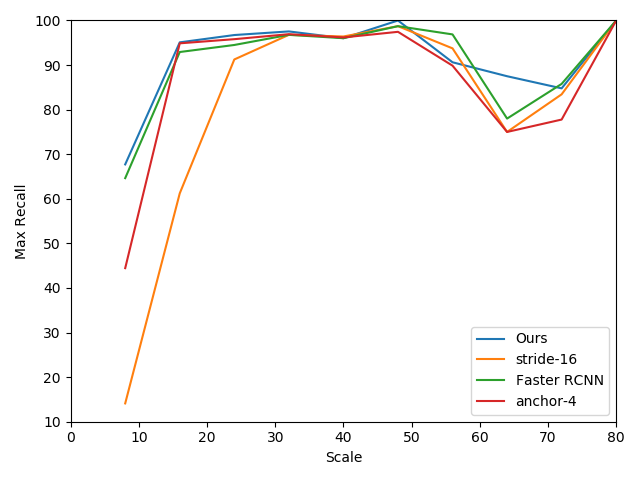}
	\end{center}
	\caption{\emph{Max-Recall} of different scales in \emph{Gray-test-dev}.}
	\label{fig.scales}
\end{figure}

\subsubsection{\bf \emph{Strong}}
To validate the effectiveness of $\mathcal{R}^2$-CNN on tiny object detection, we evaluate our network on \emph{Rgb-test-dev} and \emph{Gray-test-dev}. The results are shown in Table~\ref{tab:det-1} and Table~\ref{tab:det-2}.

\textbf{ \emph{How important is the zoom-out and zoom-in architecture?} }
Considering the large memory needed by ResNet-50 with an anchor-stride of 8 pixels, a detector with ResNet-50 is implemented with an anchor-stride of 16 pixels. The results of \emph{stride-16} with $\mathcal{R}^2$-CNN only get 83.68 mAP in \emph{Gray-test-dev}.
Besides, the mAP of different scales is shown in Fig~\ref{fig.scales}.
The performance of objects larger than 32 pixels is basically comparable to our method,
but the results of small objects drop obviously without the architecture.
In addition, we attempt to attach the global attention block to \emph{conv-4} directly. The results of \emph{No Conv-5} drop 2 points compared to our method.
This modification shows that \emph{Conv-5} can extract more high-level features that are benefited for object detection, proving the effectiveness of this architecture.

\begin{table}
  \renewcommand\arraystretch{1.5}
  \begin{center}
    \caption{Comparison experiments with FPN Faster R-CNN and Mask R-CNN on \emph{rgb-test-dev}.
     - means training without classifier.}
    \label{tab:fpn}
    \begin{tabular}{c|c|c}
      \hline
      \textbf{Metric} & mAP ($\%$) & Max Recall ($\%$) \\
      \hline
      	\textbf{FPN Faster RCNN} &  96.07 & 96.55 \\
      \hline
      	\textbf{Mask R-CNN}      &  96.20 & 96.63 \\
      \hline
      	\textbf{FPN Faster R-CNN with Tiny-Net} &  95.37 & 96.28 \\
      \hline
        \textbf{Ours$^{-}$}      &  95.81 & 96.04 \\
      \hline
        \textbf{$\mathcal{R}^2$-CNN} &  96.04 & 97.26 \\
      \hline
    \end{tabular}
  \end{center}
\end{table}

\begin{figure*}[t]
	\begin{center}
		\includegraphics[width=1.0\linewidth]{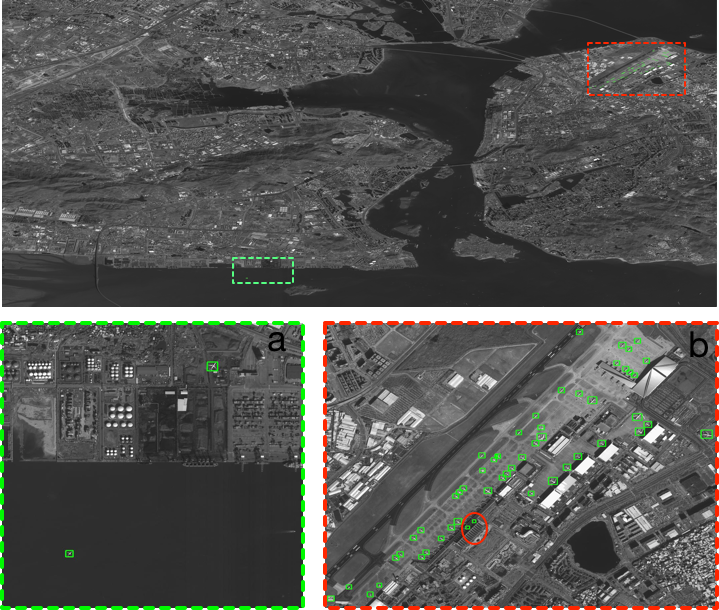}
	\end{center}
	\caption{Results of our $\mathcal{R}^2$-CNN with a score threshold of 0.9. Two airplanes are flying back to airport in image-a. Results of the airport is shown in image-b. Two objects marked by the red circle in image-b are false positives.}
	\label{fig.6}
\end{figure*}

\textbf{\emph{How important is our anchor strategy?} }
The number of clustered points is a handcraft parameter in k-means. We attempt to cluster anchor scales with a number of 4, 5, 6 and 8. The results with different anchor scales are shown in the following:

4 anchor scales: $[2.87, 5.44, 11.39, 26.72]$

5 anchor scales: $[2.32, 4.51, 7.62, 13.57, 28.34]$

6 anchor scales: $[2.18, 3.90, 6.17, 9.8, 15.91, 29.83]$

8 anchor scales: $[1.97, 3.14, 4.8, 7.05, 10.53, 15.52, 24.55]$

\emph{Max-Recall} of different scales is shown in Table~\ref{tab:det-1} and Table~\ref{tab:det-2}.
The results show that the distribution of anchors can better fit the data set to reach better performance with this strategy.
We found that \emph{Anchor-6} gets an excellent trade-off between efficiency and performance, so that it is the final parameter of our $\mathcal{R}^2$-CNN.

\textbf{ \emph{How important is position-sensitive RoI pooling?}}
Compared with Faster R-CNN\cite{ren2015faster}, our $\mathcal{R}^2$-CNN improves mAP by 2.35 points in \emph{Gray-test-dev}, particularly in tiny objects.
The spatial information is better encoded via position-sensitive RoI pooling, which is beneficial to tiny object detection.
Moreover, the time costs are 49.26 ms for Faster R-CNN but 45.21ms for $\mathcal{R}^2$-CNN with position-sensitive RoI pooling.
Through sharing all proposals' weights, we instead recalculate the feature maps of all proposals of voting from the final feature maps.
This modification is also greatly helpful for efficient process especially when there are numerous patches.

\textbf{ \emph{Comparison experiments with state-of-the-art.}}
To validate the effectiveness of our architecture, the comparison experiments are implemented with FPN Faster R-CNN and Mask R-CNN.
Considering the lack of mask annotations, Mask R-CNN is only implemented with RoI Align.
Both methods are implemented with a ResNet-50 backbone.
The results are shown in Table~\ref{tab:fpn}.
The comparable results prove the effectiveness of $\mathcal{R}^2$-CNN.
We also found that the RoI Align brings little improvement to the FPN Faster R-CNN baseline.
Considering that the tiny objects are dominating our dataset, this result is reasonable because of the potential of RoI Align is not fully explored due to the small feature maps.
The result of FPN Faster R-CNN with Tiny-Net is 0.5 points lower than $\mathcal{R}^2$-CNN training without classifier reinforced, proving the effectiveness of the specific design for Tiny-Net.
All those results show that our $\mathcal{R}^2$-CNN reaches a great trade-off within efficient processing, false positives inhibiting and tiny object detection.

\begin{figure*}[p]
	\begin{center}
		\includegraphics[width=1.0\linewidth]{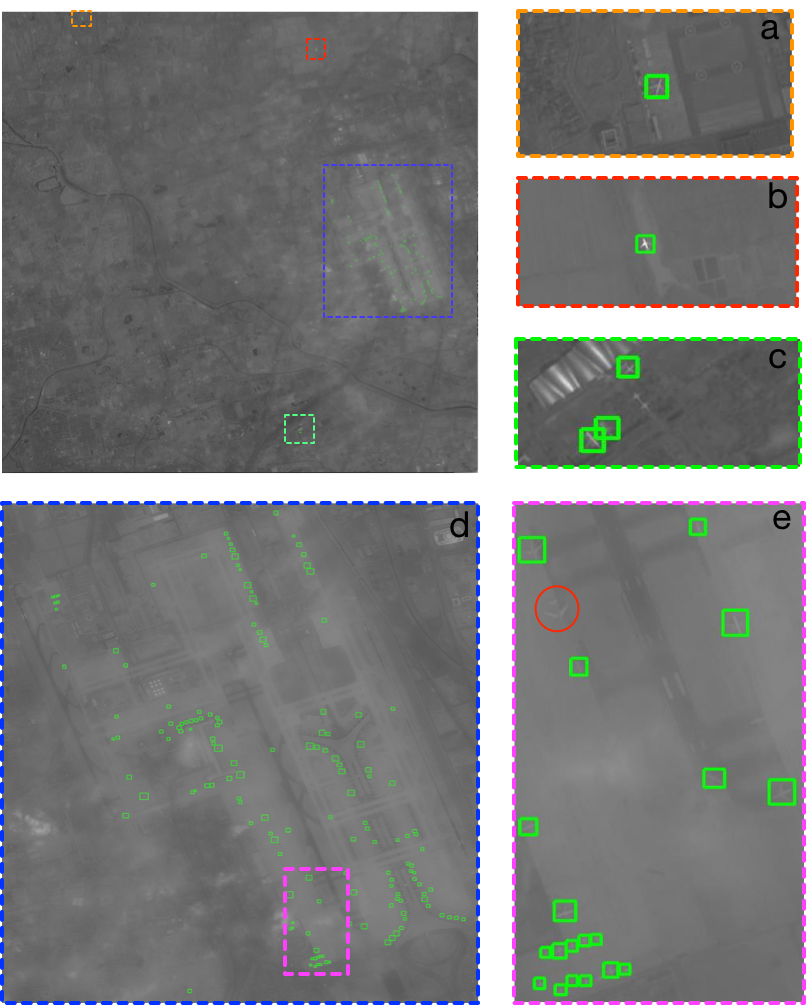}
	\end{center}
	\caption{Results of our $\mathcal{R}^2$-CNN with a score threshold of 0.9. The area is under the heavy haze. Image-a shows an airplane in the wild. Image-b shows a airplane flying in sky. Image-c shows an airport with some airplanes. Image-d shows the results of the airport. Image-e shows the details of d. The ignored airplane marked by red circle in image-e has a confidence of 0.83.}
	\label{fig.foggy}
\end{figure*}

\subsection{\bf \emph{Discussion}}
In our experiments, we get a pretty surprising results in Fig~\ref{fig.foggy}.
That is an image under the heavy haze.
We found that we can detect the aircrafts easily.
The confidence of those objects are higher than 0.9, but lower than 0.95.
It is not a high score in our results (a high confidence is larger than 0.99),
but still enough for practical engineering applications.
We must recognize that these gratifying results are not only coming from the reasonable architecture of our $\mathcal{R}^2$-CNN,
but also the precise annotation of similar situations in our training set.
The backbone, classifier and detector are specially designed to converge well while training from scratch.
That's a meaningful result for us to understand the powerful generalization ability of deep convolutional neural networks.
However, annotating all those terrible conditions very well is not a sensible selection.
But we can still explore why this is the case and how does CNN execute such well, to push the meaningful research in those situations.
There are numerous remote sensing resources to utilize and problems to solve.
With more and more powerful operators and theories, hopefully,
we can fast promote the development of real-time remote sensing systems in the future.

%% file: sections/conclusion.tex
\section{Conclusion}
We proposed $\mathcal{R}^2$-CNN, a unified and self-reinforced convolutional neural network under the end-to-end training framework, which joint the classifier and detector elegantly.
The lightweight backbone Tiny-Net extracts powerful features from the inputs quickly, and the intermediate global attention block enlarges the receptive field to inhibit false positives.
The classifier first predict the existence of detection target in the current patch,
and the specifically designed detector is followed to locate them accurately if available.
The high recall and precision in GF-1 and GF-2 validate the effectiveness of our network.
Specifically, we can process a GF-1 image in 29.4s on Titian X just with single thread.
All those experiments prove our $\mathcal{R}^2$-CNN is \emph{efficient} in both computation and memory consumption, \emph{robust} to false positives and \emph{strong} to detect tiny objects.